\documentclass[lettersize,journal]{IEEEtran}
\usepackage{amsmath,amsfonts}
\usepackage{algpseudocode}
\usepackage{algorithmicx}
\usepackage{array}
\usepackage[caption=false,font=normalsize,labelfont=sf,textfont=sf]{subfig}
\usepackage{textcomp}
\usepackage{stfloats}
\usepackage{url}
\usepackage{verbatim}
\usepackage{graphicx}
\usepackage{cite}
\usepackage{siunitx}      
\usepackage{multirow}%
\usepackage{booktabs}      
\usepackage{xcolor}%
\usepackage{graphicx}%

\newcommand{\mytablefont}{\small}      
\newcommand{\mytablearraystretch}{1.2} 
\usepackage{tikz}

\begin{document}

\title{ASSR-Net: Anisotropic Structure-Aware and Spectrally Recalibrated Network for Hyperspectral Image Fusion}

\author{Qiya Song, Hongzhi Zhou, Lishan Tan, Renwei Dian~\IEEEmembership{Member,~IEEE} and Shutao Li~\IEEEmembership{Fellow,~IEEE}
		\thanks{
        This work is supported in part by the National Natural Science Foundation of China under Grant 62401204. 
        }
		\IEEEcompsocitemizethanks{
			\IEEEcompsocthanksitem Qiya Song and Hongzhi Zhou are with School of Information Science and Engineering, Hunan Normal University, Changsha, Hunan 410081, China. 
          \IEEEcompsocthanksitem  Lisan Tan, Renwei Dian and Shutao Li are with the School of Robotics, Hunan University, Changsha, Hunan 410082, China. 
}
}

\maketitle

\begin{abstract}
Hyperspectral image fusion aims to reconstruct high-spatial-resolution hyperspectral images (HR-HSI) by integrating complementary information from multi-source inputs. Despite recent progress, existing methods still face two critical challenges: (1) inadequate reconstruction of anisotropic spatial structures, resulting in blurred details and compromised spatial quality; and (2) spectral distortion during fusion, which hinders fine-grained spectral representation. To address these issues, we propose \textbf{ASSR-Net}: an Anisotropic Structure-Aware and Spectrally Recalibrated Network for Hyperspectral Image Fusion. ASSR-Net adopts a two-stage fusion strategy comprising anisotropic structure-aware spatial enhancement (ASSE) and hierarchical prior-guided spectral calibration (HPSC). In the first stage, a directional perception fusion module adaptively captures structural features along multiple orientations, effectively reconstructing anisotropic spatial patterns. In the second stage, a spectral recalibration module leverages the original low-resolution HSI as a spectral prior to explicitly correct spectral deviations in the fused results, thereby enhancing spectral fidelity. Extensive experiments on various benchmark datasets demonstrate that ASSR-Net consistently outperforms state-of-the-art methods, achieving superior spatial detail preservation and spectral consistency.
\end{abstract}


\section{Introduction}
\IEEEPARstart{H}{yperspectral} imaging acquires fine-grained spectral information across hundreds of narrow, contiguous bands. This capability provides distinct advantages for precise material identification and quantitative analysis, rendering it indispensable in applications such as environmental monitoring~\cite{enviormental1}, precision agriculture~\cite{CoreIdentification,precision1}, mineral exploration~\cite{MP1}, and defense reconnaissance~\cite{Reconnaissance}. Nevertheless, physical limitations of imaging sensors introduce a fundamental trade-off between spatial and spectral resolution~\cite{review}, which significantly constrains the practical utility of hyperspectral imaging in scenarios that require high spatial detail.\par
To address this limitation, hyperspectral and multispectral fusion imaging has emerged as a promising computational strategy. Its objective is to reconstruct high-spatial-resolution hyperspectral images (HR-HSI) by integrating complementary information from multiple sources. Early fusion methods predominantly relied on traditional approaches, including component substitution~\cite{CS1}, multi-resolution analysis~\cite{MRA1,MRA2}, and linear models based on matrix factorization~\cite{MF1, LIU2025130777} and tensor decomposition~\cite{TD1,TD}. While these methods offer theoretical interpretability, their linear assumptions and limited capacity for nonlinear modeling often result in spatial blurring and spectral distortion~\cite{Unsupervised,10115230}. The advent of deep learning has brought substantial advancements to this field. Convolutional neural networks (CNNs)~\cite{HSRNET, DHPDARN} and Transformer architectures~\cite{LRTN, DSPNet, song2026s} have demonstrated superior ability to capture complex spatial-spectral relationships. More recently, attention-based fusion networks~\cite{MIMO-SST} and state-space models~\cite{FusionMamba} have introduced new paradigms for modeling intricate dependencies in hyperspectral data, effectively mitigating some limitations of traditional methods through enhanced nonlinear representation.\par
\begin{figure}
    \centering
    \includegraphics[width=\linewidth]{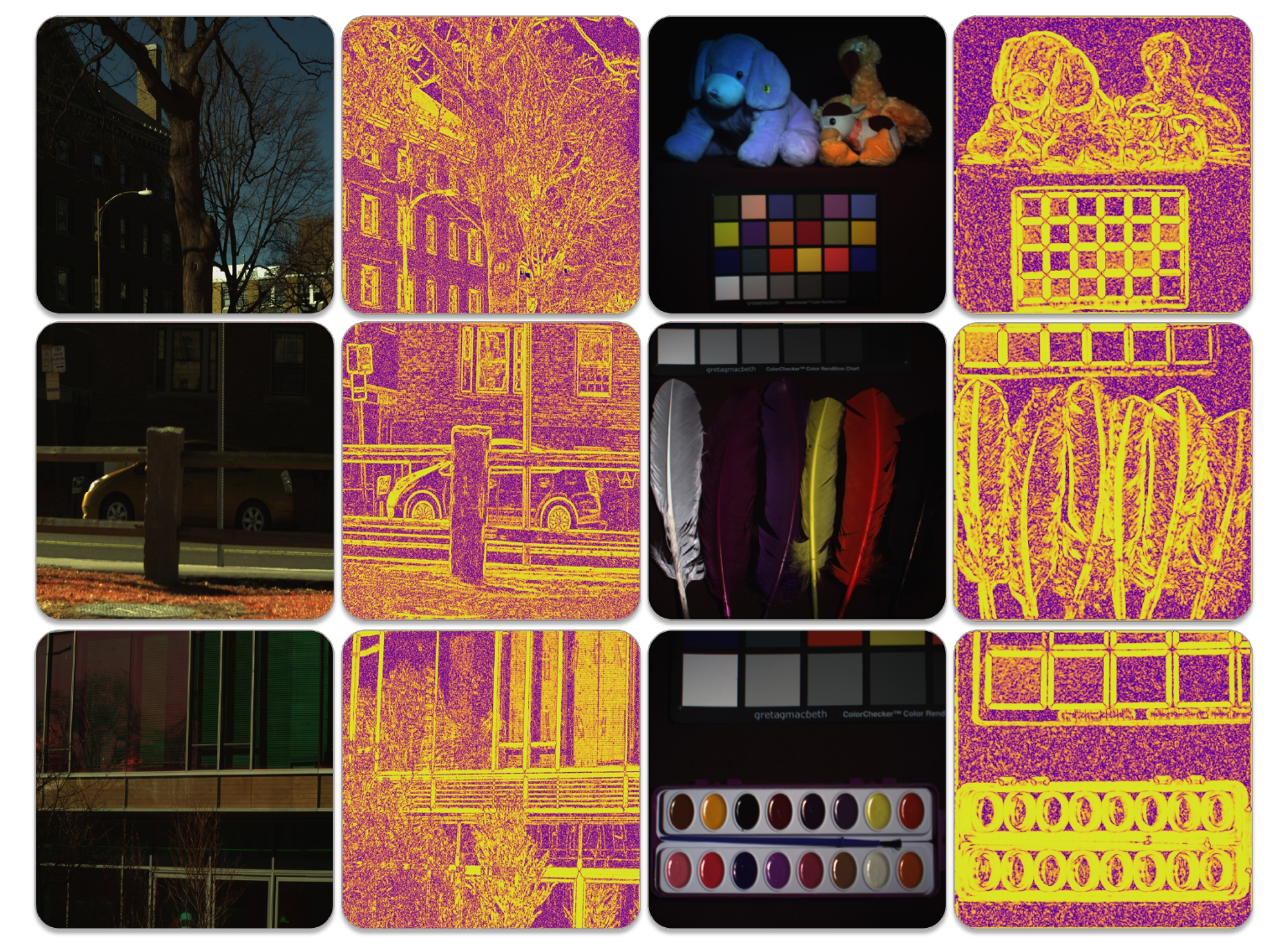}
    \caption{Anisotropy response map of the input image. Brighter regions indicate stronger anisotropy, demonstrating that the image contains substantial directional features.}
    \label{fig:placeholder}
    \vspace{-10pt}
\end{figure}
Despite these advances, contemporary deep learning methods still encounter challenges in modeling complex spatial-spectral characteristics. Spatially, standard CNNs~\cite{HSRNET,DHPDARN,SLRCNN} employ isotropic kernels, which are insufficient for capturing anisotropic structures such as edges and boundaries. Subsequent methods~\cite{FCformer,LRTN,song2025mcfnet,10536168} have incorporated directional encoding or multi-scale designs to address this issue. Nevertheless, they remain inadequate in modeling orientation-dependent features across multiple scales, often producing blurred reconstructions of linear structures. As illustrated in Fig.~\ref{fig:placeholder}, the input hyperspectral image exhibits pronounced anisotropic features, underscoring the necessity of explicitly modeling directional structures~\cite{10298274,10713289,10298249}. In addition to spatial limitations, maintaining spectral fidelity remains a critical challenge.   Existing methods~\cite{MCT-Net,SSMamba,10566044} typically rely on implicit feature learning or indirect constraints to preserve spectral consistency, lacking explicit mechanisms to anchor the reconstructed spectra to the original LR-HSI.  These limitations in directional representation and spectral fidelity fundamentally constrain the performance of existing hyperspectral image fusion methods. Furthermore, a fundamental limitation of existing single-stage fusion methods arises from the inherent conflict between spatial enhancement and spectral fidelity. In conventional single-stage approaches, joint spatial-spectral optimization is performed within a unified network. During this process, the broadband spectral characteristics of the MSI inevitably interfere with the narrowband spectral signatures of the HSI, thereby impeding fine-grained spectral representation.\par
To overcome these challenges, we propose a novel Anisotropic Structure-Aware and Spectrally Recalibrated Network (ASSR-Net). Our method employs a dual-stage fusion strategy that progressively enhances spatial structures and optimizes spectral calibration. In the Anisotropic Structure-Aware Spatial Enhancement (ASSE) stage, an Variable Direction-Aware Encoders (VDAE) module captures anisotropic features through a multi-scale geometric transformation framework, integrating multi-scale subband decomposition with directional feature extraction to effectively reconstruct detailed spatial patterns. In the Hierarchical Prior-Guided Spectral Calibration (HPSC) stage, a Global Spectral Recalibration Transformer (GSRT) mitigates spectral distortions via hierarchical spectral prior integration. This module establishes a spectral-guided attention mechanism that dynamically adjusts feature representations based on low-resolution HSI characteristics. By incorporating multi-scale spectral constraints, our network maintains spectral consistency while enhancing spatial resolution.
The two-stage architecture of ASSR-Net is explicitly designed to address spectral contamination by decoupling spatial enhancement from spectral calibration. (1) correcting spectral distortions within an already spatially coherent structure is more effective than simultaneous spatial-spectral optimization in a single stage; (2) spectral contamination can be proactively mitigated by first establishing a spatially plausible foundation, even if minor spectral deviations are introduced, followed by targeted spectral calibration; (3) a decoupled design allows each stage to be specialized and optimized for its specific objective (spatial detail injection versus spectral fidelity restoration) without forcing a compromise between these conflicting goals within a single network.
\begin{figure*}[t]
\centering
\includegraphics[width=\linewidth]{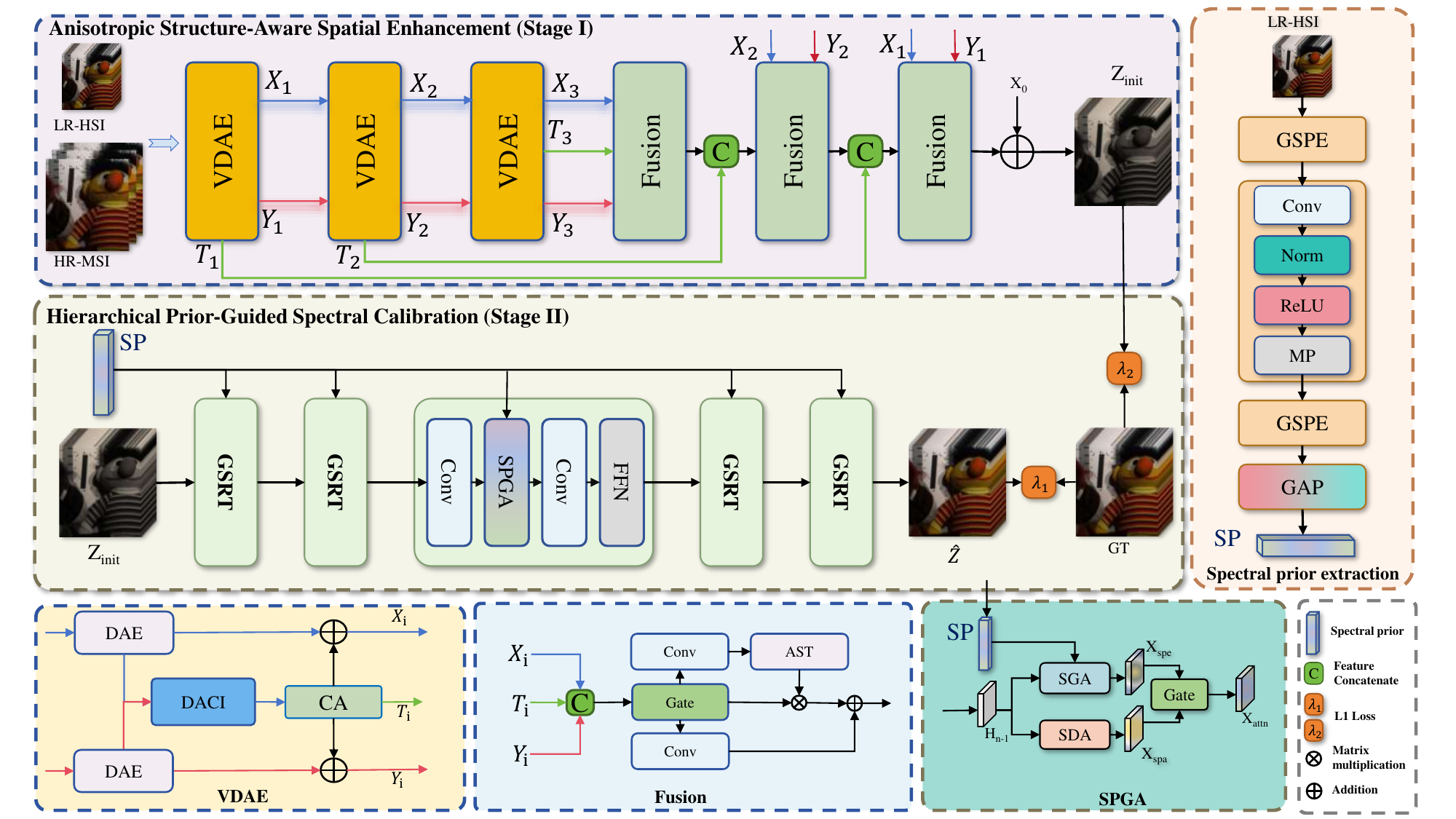}
\caption{Architecture of the proposed ASSR-Net. Stage I enhances anisotropic spatial structures from the low-resolution hyperspectral image (LR-HSI) $\mathbf{X}$ and high-resolution multispectral image (HR-MSI) $\mathbf{Y}$ using a cascade of Variable Direction-Aware Encoder (VDAE), producing an intermediate estimate $\mathbf{Z}_{\text{init}}$. Stage II then refines spectral fidelity via a Global Spectral Recalibration Transformer (GSRT) that incorporates a spectral prior extracted from $\mathbf{LR-HSI}$, yielding the Reconstructed HR-HSI $\hat{\mathbf{Z}}$. Two L1 losses, $\lambda_1$ (for $\hat{\mathbf{Z}}$) and $\lambda_2$ (for $\mathbf{Z}_{\text{init}}$), jointly supervise training.}
\label{fig:overview}
\vspace{-10pt}
\end{figure*}
The main contributions of this work are summarized as follows:
\begin{itemize}
\item We propose ASSR-Net, an end-to-end dual-stage fusion network that progressively refines spatial structures and spectral fidelity to achieve high-quality hyperspectral image reconstruction.
\item We design an VDAE module that effectively captures anisotropic spatial structures through geometric transformation and multi-scale directional analysis, significantly improving the reconstruction of detailed spatial patterns.
\item We develop a GSRT module that preserves spectral characteristics through hierarchical prior guidance and cross-scale attention mechanisms, effectively reducing spectral distortion in heterogeneous regions.
\end{itemize}
\section{Related Work}
\label{sec:related}
\subsection{Traditional Methods}
Traditional HSI-MSI fusion techniques depend on manually designed priors and linear modeling assumptions. Among them, the component substitution (CS) methods~\cite{CS1} replace certain components of the LR-HSI with spatial information from the HR-MSI. 
While these approaches are computationally efficient, they frequently introduce significant spectral distortion arising from mismatches between the substituted components and the original spectral characteristics of the LR-HSI.
Multiresolution analysis (MRA) methods~\cite{MRA1, MRA2} enhance spatial resolution by injecting high-frequency details derived from multi-scale decomposition of the HR-MSI. Although MRA methods generally achieve superior spectral preservation compared to CS approaches, their reliance on isotropic filters inherently limits their capacity to reconstruct anisotropic or directional structures. 
Matrix factorization~\cite{MF1, mp2, MF} and tensor decomposition~\cite{TD} methods represent the HSI data via low-rank factorized models to capture inherent spatial-spectral correlations. However, their underlying linearity assumptions prove inadequate for modeling the complex, nonlinear relationships inherent between the LR-HSI and HR-MSI data.
To overcome these limitations, recent tensor-based approaches have introduced more sophisticated regularization and learning strategies. For instance, Dian et al. proposed a generalized tensor nuclear norm regularization that flexibly exploits the low-rank structure of hyperspectral data~\cite{TD1}. Wang et al. developed an unsupervised deep Tucker decomposition network integrating spatial-spectral manifold learning for blind fusion~\cite{Unsupervised}.

\subsection{Deep Learning-Based Methods}
Deep learning has significantly advanced HSI-MSI fusion, yet prevailing approaches continue to face two interconnected challenges: effectively preserving directional structures and maintaining accurate spectral fidelity. 
In spatial reconstruction, CNNs like HSRNet~\cite{HSRNET} establish mappings through local feature extraction, but their isotropic kernels treat all spatial directions uniformly, often blurring linear features and edges. While Transformer architectures~\cite{LRTN,FCformer} overcome limited receptive fields via self-attention and capture long-range dependencies, their uniform attention weights still fail to emphasize dominant orientations. To address these limitations, multi-scale and multi-stage architectures have been explored. Dong et al. proposed a feature pyramid fusion network that aggregates multi-resolution representations for hyperspectral pansharpening~\cite{10298274}. Wu et al. designed a multistage spatial-spectral fusion network that cascades multiple U-shaped sub-networks for spectral super-resolution~\cite{10713289}. These designs show the benefit of progressive refinement, yet most still stack functionally similar blocks.
More recent innovations, including state-space models such as FusionMamba~\cite{FusionMamba} and diffusion models~\cite{PLRDiff}, offer efficient global modeling and generate high-frequency details. Despite these strengths, their sequential scanning or stochastic denoising mechanisms remain non-adaptive to anisotropic structures, often prioritizing general detail over directional accuracy. 
In spectral fidelity, researchers have employed spectral attention~\cite{SpecFormer} and selective re-learning~\cite{SRLF} to model inter-band relationships and address distortions. However, these methods primarily rely on indirect constraints or static physical priors~\cite{LRTN}, lacking explicit mechanisms to directly anchor the output spectrum to the high-fidelity reference of the original LR-HSI. This limitation restricts dynamic spectral calibration during reconstruction, particularly in heterogeneous regions with mixed materials, leading to persistent spectral distortion.
\section{Proposed Method}
\subsection{Overview}
\label{sec:overview}
The proposed ASSR-Net adopts a dual-stage architecture that decouples spatial enhancement from spectral calibration. As illustrated in Fig.~\ref{fig:overview}, the network takes a low-resolution hyperspectral image (LR-HSI) $\mathbf{X} \in \mathbb{R}^{C \times H \times W}$ and a high-resolution multispectral image (HR-MSI) $\mathbf{Y} \in \mathbb{R}^{c \times H \times W}$ as inputs, and reconstructs a high-resolution hyperspectral image $\hat{\mathbf{Z}} \in \mathbb{R}^{C \times H \times W}$.The first stage performs anisotropic spatial enhancement, generating an initial estimate $\mathbf{Z}_{\text{init}}$ that inherits high-frequency details from $\mathbf{Y}$ while preserving the spectral structure of $\mathbf{X}$. The second stage refines the spectral fidelity by using the original LR-HSI $\mathbf{X}$ as a spectral prior. The overall computation can be expressed as:
\begin{equation}
\begin{split}
&\mathbf{Z}_{\text{init}} = \text{ASSE}(\mathbf{X},\mathbf{Y}),\\
&\hat{\mathbf{Z}} = \text{HPSC}(\mathbf{Z}_{\text{init}}, \mathbf{X}).
\end{split}
\end{equation}

\subsection{Stage I: Anisotropic Structure-Aware Spatial Enhancement (ASSE)}
\label{sec:asaf}
ASSE aims to produce an initial high-resolution estimate $\mathbf{Z}_{\text{init}}$ with rich spatial details. It first pre-processes the inputs:
\begin{equation}
\begin{split}
&\mathbf{X}_0 = \text{UpSample}(\mathbf{X}),\\
&\mathbf{Y}_0 = \text{Conv}_{3\times3}(\mathbf{Y}),
\end{split}
\end{equation}
where $\text{UpSample}$ is bilinear upsampling to match the spatial dimensions of $\mathbf{Y}$. The core of ASSE is a cascade of three Variable Direction-Aware Encoder (VDAE), each of which refines the features while extracting directional cross-modal correspondences:
\begin{equation}
(\mathbf{X}_k,\mathbf{Y}_k,\mathbf{T}_k) = \text{VDAE}_k(\mathbf{X}_{k-1},\mathbf{Y}_{k-1}),\quad k=1,2,3.
\end{equation}
After the three encoding stages, the multi-scale directional tensors $\mathbf{T}_k$ and the features $\mathbf{X}_k,\mathbf{Y}_k$ are progressively fused through three learnable fusion modules:
\begin{equation}
\begin{split}
&\mathbf{F}_1 = \text{Fusion}_1(\mathbf{T}_3,\mathbf{X}_3,\mathbf{Y}_3),\\
&\mathbf{F}_2 = \text{Fusion}_2([\mathbf{F}_1,\mathbf{T}_2],\mathbf{X}_2,\mathbf{Y}_2),\\
&\mathbf{F}_3 = \text{Fusion}_3([\mathbf{F}_2,\mathbf{T}_1],\mathbf{X}_1,\mathbf{Y}_1),
\end{split}
\end{equation}
where $[\cdot,\cdot]$ denotes channel-wise concatenation. Finally, the initial estimate is obtained by a residual addition:
\begin{equation}
\mathbf{Z}_{\text{init}} = \mathbf{X}_0 + \mathbf{F}_3.
\end{equation}
This residual formulation ensures that the network learns to predict the high-frequency residual details $\mathbf{F}_3$ rather than the full image, thereby stabilizing training and preserving the spectral structure of the upsampled LR-HSI $\mathbf{X}_0$.

\begin{figure}[t]
\centering
\includegraphics[width=\columnwidth]{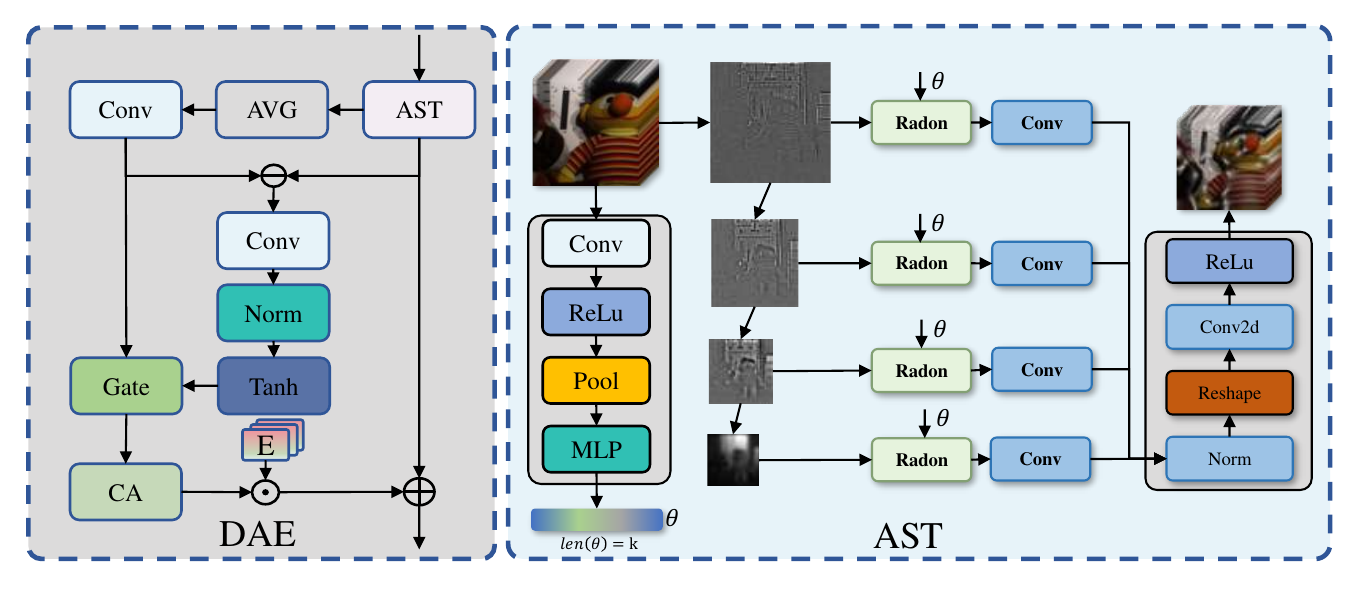}
\caption{Detailed architecture of the Directional Attention Enhancement (DAE) module and its core component, the Anisotropic Structure Transform (AST). DAE enhances directional features via a global-local decomposition followed by a learnable gating fusion. AST performs multi-scale subband decomposition, differentiable Radon approximation along adaptively predicted directions, and frequency-adaptive enhancement in the wavelet domain to capture anisotropic structures.}
\label{fig:vdae}
\vspace{-10pt}
\end{figure}

\textbf{Variable Direction-Aware Encoder (VDAE).}
As illustrated in Fig.~\ref{fig:overview}, each VDAE consists of three steps. First, a Directional Attention Enhancement (DAE) module independently enhances anisotropic structures:
\begin{equation}
\begin{split}
&\mathbf{F}_x^{(k)} = \text{DAE}^{(k)}(\mathbf{X}_{k-1}),\\
&\mathbf{F}_y^{(k)} = \text{DAE}^{(k)}(\mathbf{Y}_{k-1}).
\end{split}
\end{equation}

Second, a Dual-pathway Adaptive Cross Interaction (DACI) module establishes cross-modal directional correspondences:
\begin{equation}
\mathbf{T}_k = \text{DACI}^{(k)}(\mathbf{F}_x^{(k)},\mathbf{F}_y^{(k)}).
\end{equation}
Here, $\mathbf{T}_k$ serves as a directional correlation tensor that encodes how anisotropic structures (edges, textures) in the high-resolution MSI $\mathbf{Y}$ should guide the spatial enhancement of the low-resolution HSI $\mathbf{X}$, enabling cross-modal information transfer while maintaining modality-specific characteristics.

Third, $\mathbf{T}_k$ is refined by a channel attention and used to update the features:
\begin{equation}
\begin{split}
&\mathbf{T}_k^{\text{ref}} = \mathbf{T}_k \odot \sigma\big(\text{MLP}(\text{GAP}(\mathbf{T}_k))\big),\\
&\mathbf{X}_k = \mathbf{F}_x^{(k)} + \text{Conv}_X^{(k)}(\mathbf{T}_k^{\text{ref}}),\\
&\mathbf{Y}_k = \mathbf{F}_y^{(k)} + \text{Conv}_Y^{(k)}(\mathbf{T}_k^{\text{ref}}),
\end{split}
\end{equation}
where $\text{Conv}_X^{(k)},\text{Conv}_Y^{(k)}$ are $1\times1$ convolutions. The channel attention mechanism (via GAP and MLP) adaptively recalibrates the importance of different directional features based on global context, suppressing irrelevant orientations while emphasizing dominant structural directions present in the scene.

\textbf{Directional Attention Enhancement (DAE).}
DAE enhances anisotropic structures via the Anisotropic Structure Transform (AST), as illustrated in Fig.~\ref{fig:vdae}. The overall DAE process is:
\begin{equation}
\begin{split}
&\mathbf{F}_{\text{directional}} = \text{AST}(\mathbf{F}),\\
&\mathbf{F}_{\text{global}} = \text{GlobalPath}(\mathbf{F}_{\text{directional}}),\\
&\mathbf{F}_{\text{local}} = \text{LocalPath}(\mathbf{F}_{\text{directional}} - \mathbf{F}_{\text{global}}),\\
&\text{DAE}(\mathbf{F}) = \mathcal{G}(\mathbf{F}_{\text{global}}, \mathbf{F}_{\text{local}}) + \mathbf{F},
\end{split}
\end{equation}
where $\text{GlobalPath}$ captures context via average pooling and pointwise convolution, $\text{LocalPath}$ enhances fine details via two $3\times3$ convolutions with instance normalization and Tanh activation, and $\mathcal{G}$ is a learnable gate. This global-local decomposition follows the classical image processing paradigm of separating low-frequency structure from high-frequency detail. The learnable gate $\mathcal{G}$ dynamically balances these components based on local image statistics, ensuring that directional features are enhanced without over-amplifying noise.

\textbf{Anisotropic Structure Transform (AST).}
AST performs multi-scale directional analysis to produce a feature map rich in orientation information. It decomposes the input, extracts directional responses along adaptively predicted angles, and enhances them in the frequency domain.

In principle, the ideal way to capture linear structures aligned with a direction $\theta$ is the Radon transform, which computes line integrals:
\begin{equation}
\mathcal{R}[\mathbf{D}_k](\theta_i, \rho) = \iint \mathbf{D}_k(x,y)\,\delta(x\cos\theta_i+y\sin\theta_i-\rho)\,dx\,dy.
\label{eq:radon}
\end{equation}
This transform maps the image space to a projection space, where each point $(\theta_i, \rho)$ represents the integral intensity along a line at angle $\theta_i$ and distance $\rho$ from the origin. Linear structures aligned with $\theta_i$ produce distinctive peak responses, enabling explicit detection of directional patterns regardless of their spatial position.

However, this operator is non-differentiable, making it unsuitable for end-to-end gradient-based learning. To overcome this limitation, AST employs a differentiable approximation $\widetilde{\mathcal{R}}[\cdot]$ that replaces the exact line integral with a combination of coordinate rotation, grid sampling, and average pooling.

The complete AST operation is:
\begin{equation}
\mathbf{F}_{\text{directional}} = \Psi\Big( \big\{ \mathbf{D}_k^{\text{enhanced}} \big\}_{k=1}^{K}, \mathbf{C} \Big),
\end{equation}
with
\begin{equation}
\mathbf{D}_k^{\text{enhanced}} = \mathbf{D}_k + \alpha_k \cdot \mathcal{W}^{-1}\big[\mathcal{W}[\widetilde{\mathcal{R}}[\mathbf{D}_k]] \odot \mathcal{M}_k\big],
\label{eq:enhancement}
\end{equation}
where $\{\mathbf{D}_k\}_{k=1}^{4}$ are detail subbands at different scales, the coarse approximation is $\mathbf{C} = \mathbf{D}_4$, $\alpha_k$ is a learnable scalar, $\mathcal{W}$ and $\mathcal{W}^{-1}$ denote the Haar wavelet transform and its inverse, $\mathcal{M}_k$ is a learnable frequency modulation mask, $\widetilde{\mathcal{R}}[\cdot]$ is the differentiable Radon approximation, and $\Psi$ reconstructs the directional feature map via upsampling and learned convolutions. Equation~\eqref{eq:enhancement} performs frequency-adaptive enhancement of directional features. The wavelet transform $\mathcal{W}$ decomposes the Radon projection into different frequency bands, where the learnable mask $\mathcal{M}_k$ selectively amplifies frequency components corresponding to significant directional structures while suppressing noise, thereby implementing a data-driven multi-scale anisotropic filter. The individual steps are defined as follows:\par
\textbf{Multi-scale Subband Decomposition.} Starting from the input feature $\mathbf{F} \in \mathbb{R}^{C \times H \times W}$, set $\mathbf{F}_1 = \mathbf{F}$. For $k = 1,2,3,4$:
\begin{equation}
\begin{split}
&\mathbf{D}_k = \mathbf{F}_k \ast_x \mathbf{g}_{s_k}^x \ast_y \mathbf{g}_{s_k}^y,\\
&\mathbf{F}_{k+1} = \text{AvgPool}_{2\times2}(\mathbf{D}_k),
\end{split}
\end{equation}
where $\ast_x$ and $\ast_y$ are 1D convolutions along the horizontal and vertical axes with Gaussian kernels $\mathbf{g}_{s_k}^x,\mathbf{g}_{s_k}^y$ of sizes $(7,5,3,3)$. $\mathbf{F}_k$ is the feature map at the $k$-th scale, $\mathbf{D}_k$ is the detail subband at scale $k$, and $\text{AvgPool}_{2\times2}$ is average pooling with stride 2. The coarse approximation is $\mathbf{C} = \mathbf{D}_4$. This Laplacian-style pyramid decomposition separates the feature map into band-pass detail layers $\mathbf{D}_k$ (capturing structures at specific scales) and a low-pass residual $\mathbf{C}$. The Gaussian smoothing ensures that each scale captures directional features at a specific spatial frequency range, enabling scale-specific anisotropic analysis.

\textbf{Adaptive Direction Prediction.} Instead of using fixed projection angles, AST predicts $K$ projection angles $\boldsymbol{\theta}=[\theta_1,\dots,\theta_K]\in[0,\pi)^K$ via a lightweight network:
\begin{equation}
\begin{split}
&\mathbf{H}_1 = \text{ReLU}(\text{Conv}_{3\times3}(\mathbf{F})),\\
&\mathbf{H}_2 = \text{ReLU}(\text{Conv}_{3\times3}(\mathbf{H}_1)),\\
&\mathbf{h}_{\text{pool}} = \text{AdaptiveAvgPool2d}_{(8,8)}(\mathbf{H}_2),\\
&\mathbf{h}_{\text{flat}} = \text{Flatten}(\mathbf{h}_{\text{pool}}),\\
&\mathbf{h}_{\text{fc1}} = \text{ReLU}(\mathbf{W}_1 \mathbf{h}_{\text{flat}} + \mathbf{b}_1),\\
&\boldsymbol{\theta} = \pi \cdot \text{Sigmoid}(\mathbf{W}_2 \mathbf{h}_{\text{fc1}} + \mathbf{b}_2).
\end{split}
\end{equation}
Rather than using fixed orientations , this subnetwork analyzes the global feature statistics to predict the most relevant projection angles $\boldsymbol{\theta}$ for the current image. This data-adaptive approach ensures that computational resources are focused on the dominant structural directions actually present in the scene, improving both efficiency and accuracy.

\textbf{Differentiable Radon approximation.} For each detail subband $\mathbf{D}_k$ and each predicted direction $\theta_i$, we approximate the Radon transform using coordinate rotation, grid sampling, and average projection:
\begin{equation}
\begin{split}
&\begin{bmatrix} x' \\ y' \end{bmatrix} =
\begin{bmatrix} \cos\theta_i & \sin\theta_i \\ -\sin\theta_i & \cos\theta_i \end{bmatrix}
\begin{bmatrix} x \\ y \end{bmatrix},\\
&\mathbf{R}_i = \text{GridSample}(\mathbf{D}_k, \mathbf{G}_{\text{rot}}),\\
&\mathbf{sp}_{\theta_i}^{(k)} = \frac{1}{H}\sum_{h=1}^{H} \mathbf{R}_i[:, h, :],
\end{split}
\end{equation}
where $\mathbf{G}_{\text{rot}}$ is the rotated coordinate grid normalized to $[-1,1]$, and $\text{GridSample}$ performs bilinear interpolation. The complete approximation is:
\begin{equation}
\widetilde{\mathcal{R}}[\mathbf{D}_k] = \bigoplus_{i=1}^{K} \mathbf{sp}_{\theta_i}^{(k)},
\end{equation}
with $\bigoplus$ denoting concatenation along the channel dimension. This approximation implements the Radon transform through differentiable operations: rotation aligns the image with the projection axis, grid sampling resamples the rotated image, and average pooling along the vertical axis computes the line integral. The concatenation stacks projections from all $K$ directions, creating a multi-orientation feature representation that encodes the strength of linear structures at each angle.

\textbf{Frequency-adaptive Enhancement.} The Radon representation is enhanced in the wavelet domain:
\begin{equation}
\mathbf{D}_k^{\text{enhanced}} = \mathbf{D}_k + \alpha_k \cdot \mathcal{W}^{-1}\big[\mathcal{W}[\widetilde{\mathcal{R}}[\mathbf{D}_k]] \odot \mathcal{M}_k\big].
\end{equation}
Here, $\mathcal{W}[\widetilde{\mathcal{R}}[\mathbf{D}_k]]$ transforms the directional projection into wavelet coefficients, where the learnable mask $\mathcal{M}_k$ performs element-wise modulation. This allows the network to selectively enhance or suppress specific frequency components of the directional response, effectively implementing a data-driven directional filter that emphasizes salient structures while attenuating noise.
The enhanced subbands and coarse component are aggregated via $\Psi$ to produce the final directional feature map $\mathbf{F}_{\text{directional}}$.

\textbf{Dual-pathway Adaptive Cross Interaction (DACI).}
DACI injects HR-MSI directional details into the LR-HSI spectral representation by operating at multiple scales. It progressively downsamples the features, concatenates corresponding scale representations, modulates them with channel attention, and upsamples back. Formally,
\begin{equation}
\mathbf{T}_k = \sum_{l=1}^L \gamma_l \cdot \mathcal{A}^{(l)}\big( \mathcal{D}^{(l)}(\mathbf{F}_x^{(k)}), \mathcal{D}^{(l)}(\mathbf{F}_y^{(k)}) \big),
\end{equation}
where $\mathcal{A}^{(l)}(\cdot)$ implements cross-attention at scale $l$, $\mathcal{D}^{(l)}(\cdot)$ performs spatial downsampling to level $l$, and $\gamma_l$ are learnable fusion coefficients that balance multi-scale contributions. This multi-scale cross-attention mechanism enables fine-grained spatial-spectral alignment: at each scale $l$, the attention $\mathcal{A}^{(l)}$ determines which high-resolution MSI features should guide the enhancement of low-resolution HSI features. The learnable coefficients $\gamma_l$ adaptively weight the contribution of each scale, ensuring that fine details (small $l$) and contextual structures (large $l$) are appropriately balanced in the final fusion.

\begin{figure}[t]
\centering
\includegraphics[width=\linewidth]{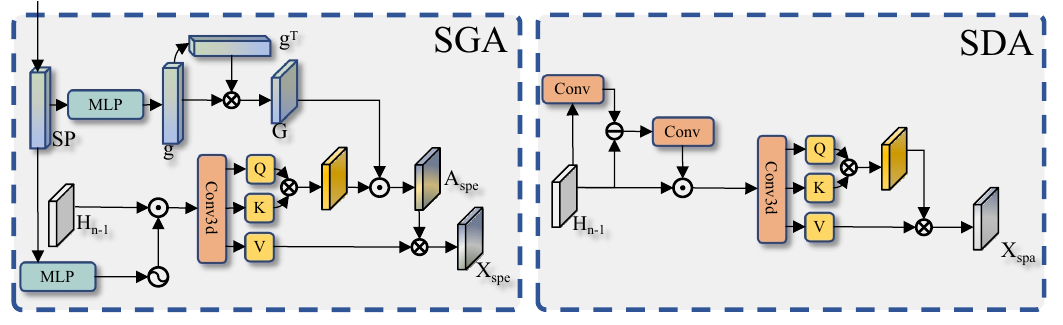}
\caption{Illustration of the two complementary attention mechanisms in the Spectral Prior-Guided Attention (SPGA) module. Spectral-Guided Attention (SGA): the spectral prior $\mathbf{sp}$ generates a channel modulation vector $\mathbf{s}$ and a guidance matrix $\mathbf{G}$, biasing self-attention weights toward spectrally consistent regions. Spatial Differential Attention (SDA): a spatial mask $\mathbf{M}_{\text{spa}}$ computed from local gradients highlights edges and textures, followed by standard self-attention. The outputs of both branches are adaptively fused via a learnable gating mechanism.}
\label{fig:spga}
\vspace{-10pt}
\end{figure}

\subsection{Stage II: Hierarchical Prior-Guided Spectral Calibration}
\label{sec:gsrt}
The core of HPSC is a Global Spectral Recalibration Transformer(GSRT), which corrects spectral distortions introduced during spatial enhancement by explicitly using the original LR-HSI $\mathbf{X}$ as a high-fidelity spectral prior. A compact spectral prior $\mathbf{sp} \in \mathbb{R}^C$ is first extracted from $\mathbf{X}$ via the Spectral Guidance (SG) module. The module is composed of $N$ cascaded transformer blocks, each processing the input $\mathbf{H}_{n-1}$ (with $\mathbf{H}_0=\mathbf{Z}_{\text{init}}$) and the spectral prior $\mathbf{sp}$.

\textbf{Spectral Prior Extraction.}
A compact spectral prior $(\mathbf{sp})\in\mathbb{R}^C$ is obtained via:
\begin{equation}
\mathbf{sp} = \mathrm{SG}(\mathbf{X}) = \mathrm{AvgPool}\big( \mathrm{Conv}_{\mathrm{down}}(\mathbf{X}) \big),
\end{equation}
where $\mathrm{Conv}_{\mathrm{down}}$ stacks three stride-2 $3\times3$ convolutions, each followed by layer normalization and ReLU, followed by global average pooling. $\mathbf{sp}$ encodes the essential spectral distribution of the original scene.

\textbf{Spectral Prior-Guided Attention (SPGA) Module.}
The SPGA module combines two complementary attention pathways and a learnable gate to produce an attention-enhanced feature $\mathbf{X}_{\text{att}}$. As illustrated in Fig.~\ref{fig:spga}, it consists of Spectral-Guided Attention (SGA) and Spatial Differential Attention (SDA), whose outputs are adaptively fused.

\textbf{Spectral-Guided Attention (SGA).}
SGA uses $\mathbf{sp}$ to modulate both feature channels and attention weights. First, a channel-wise modulation vector $\mathbf{s}$ is derived:
\begin{equation}
\begin{split}
&\mathbf{s} = \sigma(\text{MLP}(\mathbf{sp})),\\
&\mathbf{F}_{\text{mod}} = \text{LN}(\mathbf{H}_{n-1}) \odot (\mathbf{s}\cdot\mathbf{1}_{1\times H\times W}).
\end{split}
\end{equation}
A guidance matrix $\mathbf{G}=\mathbf{g}\mathbf{g}^\top$ is built from $\mathbf{g}=\text{MLP}_{\text{proj}}(\mathbf{sp})$. The self-attention then becomes:
\begin{equation}
\begin{split}
&\mathbf{Q},\mathbf{K},\mathbf{V} = \text{Proj}(\mathbf{F}_{\text{mod}}),\\
&\mathbf{A}_{\text{spe}} = \text{Softmax}\!\left( \frac{\mathbf{Q}\mathbf{K}^\top}{\sqrt{d}} \odot \mathbf{G} \right),\\
&\mathbf{X}_{\text{spe}} = \mathbf{A}_{\text{spe}}\mathbf{V}.
\end{split}
\end{equation}

\sisetup{detect-weight}
\begin{table*}[t]
\centering
\mytablefont
\setlength{\tabcolsep}{0pt}
\renewcommand\arraystretch{\mytablearraystretch}
\caption{Quantitative comparison on the CAVE and Harvard datasets. Best results are in \textbf{bold} and the second best are \underline{underlined}.}
\label{tab:comparison}
\begin{tabular*}{\textwidth}{@{\extracolsep{\fill}}
  l
  S[table-format=2.4, round-precision=4, detect-weight]
  S[table-format=1.2, round-precision=2, detect-weight]
  S[table-format=1.4, round-precision=4, detect-weight]
  S[table-format=1.4, round-precision=4, detect-weight]
  S[table-format=1.4, round-precision=4, detect-weight]
  S[table-format=2.4, round-precision=4, detect-weight]
  S[table-format=1.2, round-precision=2, detect-weight]
  S[table-format=1.4, round-precision=4, detect-weight]
  S[table-format=1.4, round-precision=4, detect-weight]
  S[table-format=1.4, round-precision=4, detect-weight]
@{}}
\toprule
\multirow{2}{*}{Method} & \multicolumn{5}{c}{CAVE Dataset} & \multicolumn{5}{c}{Harvard Dataset} \\
\cmidrule(lr){2-6} \cmidrule(lr){7-11}
 & {PSNR$\uparrow$} & {SAM$\downarrow$} & {UIQI$\uparrow$} & {SSIM$\uparrow$} & {ERGAS$\downarrow$}
 & {PSNR$\uparrow$} & {SAM$\downarrow$} & {UIQI$\uparrow$} & {SSIM$\uparrow$} & {ERGAS$\downarrow$} \\
\midrule
DHIF-Net   & 47.2893 & 2.4101 & 0.9733 & 0.9947 & 0.6034 & 47.7261 & 2.7313 & 0.8938 & \underline{0.9855} & 0.5667 \\
DSPNet     & 47.8060 & 2.5041 & 0.9729 & 0.9944 & 0.5756 & 47.6887 & 2.7325 & 0.8941 & 0.9852 & 0.5677 \\
LRTN       & 47.8958 & 2.3644 & 0.9704 & 0.9945 & 0.5612 & 47.6663 & 2.7417 & 0.8941 & 0.9852 & 0.5687 \\
MIMO-SST   & 48.2001 & 2.5618 & 0.9697 & 0.9942 & 0.5575 & 47.6364 & 2.7731 & 0.8939 & 0.9851 & 0.5697 \\
SINet      & 48.6726 & 2.3013 & 0.9743 & 0.9950 & 0.5261 & 47.8412 & 2.7110 & 0.8961 & \underline{0.9855} & 0.5627 \\
OTIAS      & 47.9541 & 2.4467 & 0.9728 & 0.9945 & 0.5646 & 47.6621 & 2.7514 & 0.8939 & 0.9851 & 0.5687 \\
SRLF       & \underline{49.2959} & \underline{2.15} & \underline{0.9749} & \underline{0.9954} & \underline{0.4924} & \underline{47.8924} & \underline{2.68} & \underline{0.8956} & \textbf{0.9856} & \underline{0.5617} \\
Ours       & \textbf{49.5820} & \textbf{2.05} & \textbf{0.9769} & \textbf{0.9961} & \textbf{0.4725} & \textbf{47.9943} & \textbf{2.66} & \textbf{0.8961} & \textbf{0.9856} & \textbf{0.5587} \\
\bottomrule
\end{tabular*}
\vspace{-10pt}
\end{table*}
\sisetup{detect-weight}
\begin{table*}[t]
\centering
\mytablefont
\setlength{\tabcolsep}{0pt}
\renewcommand\arraystretch{\mytablearraystretch}
\caption{No-reference quality assessment (QNR) on the Gaofen5 dataset. The best result is in \textbf{bold} and the second best is \underline{underlined}.}
\label{tab:qnr_results}
\begin{tabular*}{\linewidth}{@{\extracolsep{\fill}} l *{8}{c} @{}}
\toprule
Method & DHIF-Net & DSPNet & LRTN & MIMO-SST & SINet & OTIAS & SRLF & \textbf{Ours} \\
\midrule
QNR $\uparrow$ & 0.9711 & 0.9870 & 0.9862 & 0.9871 & 0.9869 & 0.9871 & \underline{0.9872} & \textbf{0.9873} \\
\bottomrule
\end{tabular*}
\vspace{-10pt}
\end{table*}
\textbf{Spatial Differential Attention (SDA).}
SDA preserves fine spatial details without using $\mathbf{sp}$. It computes a spatial mask that highlights regions with high local gradients:
\begin{equation}
\mathbf{M}_{\text{spa}} = \sigma\!\left( \text{Conv}_{1\times1}\big( \text{Conv}_{3\times3}(\text{LN}(\mathbf{H}_{n-1})) - \text{LN}(\mathbf{H}_{n-1}) \big) \right).
\end{equation}
The subtraction $\text{Conv}_{3\times3}(\cdot) - \cdot$ acts as a discrete Laplacian, emphasizing edges and textures. The subsequent $1\times1$ convolution and sigmoid produce a mask $\mathbf{M}_{\text{spa}} \in [0,1]^{1\times H\times W}$. The input is then modulated:
\begin{equation}
\mathbf{F}_{\text{mod}}^{\text{spa}} = \text{LN}(\mathbf{H}_{n-1}) \odot \mathbf{M}_{\text{spa}}.
\end{equation}
Standard self-attention is applied to $\mathbf{F}_{\text{mod}}^{\text{spa}}$:
\begin{equation}
\begin{split}
&\mathbf{Q}',\mathbf{K}',\mathbf{V}' = \text{Proj}(\mathbf{F}_{\text{mod}}^{\text{spa}}),\\
&\mathbf{A}_{\text{spa}} = \text{Softmax}\!\left( \frac{\mathbf{Q}'\mathbf{K}'^{\top}}{\sqrt{d}} \right),\\
&\mathbf{X}_{\text{spa}} = \mathbf{A}_{\text{spa}} \mathbf{V}'.
\end{split}
\end{equation}
This branch preserves high-frequency spatial information by forcing the attention to focus on locations where the gradient is strong, effectively acting as a structure-preserving regularizer.

\begin{figure*}[h]
\centering
\includegraphics[width=\textwidth]{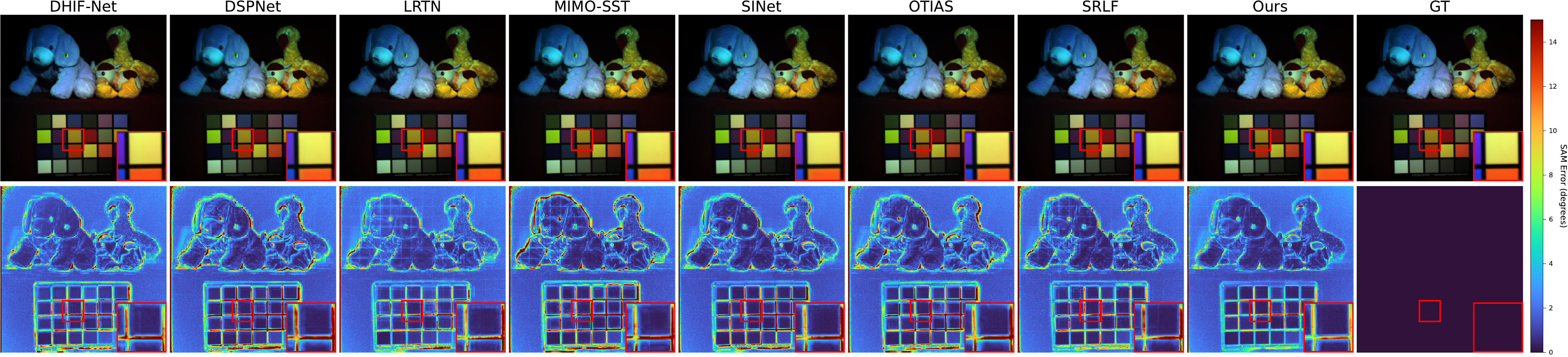}
\caption{Visual comparison on the CAVE dataset. From left to right: DHIF-Net, DSPNet, LRTN, MIMO-SST, SINet, OTIAS, SRLF, ASSR-Net (ours), and GT.}
\label{fig:compare1}
\end{figure*}
\begin{figure*}[h]
\centering
\includegraphics[width=\textwidth]{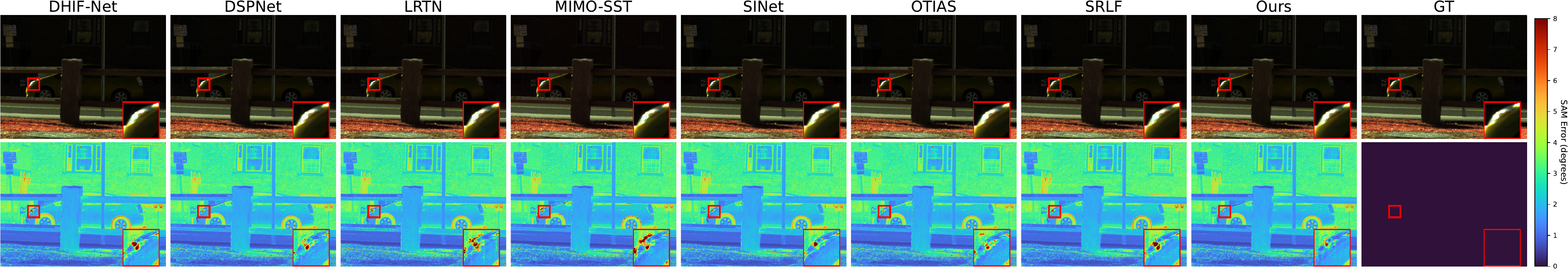}
\caption{Visual comparison on the Harvard dataset. From left to right: DHIF-Net, DSPNet, LRTN, MIMO-SST, SINet, OTIAS, SRLF, ASSR-Net (ours), and GT.}
\label{fig:compare2}
\end{figure*}
\begin{figure*}[h]
\centering
\includegraphics[width=\textwidth]{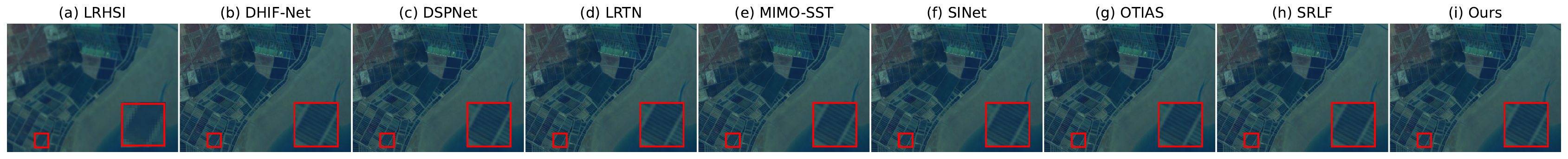}
\caption{Visual results on the Gaofen5 dataset. From left to right: DHIF-Net, DSPNet, LRTN, MIMO-SST, SINet, OTIAS, SRLF, ASSR-Net (ours), and GT.}
\label{fig:real_data}
\end{figure*}

\sisetup{detect-weight}
\begin{table*}[t]
\centering
\mytablefont
\setlength{\tabcolsep}{0pt}
\renewcommand\arraystretch{\mytablearraystretch}
\caption{Hyperparameter sensitivity analysis of the number of projection directions $K$ in ASSE and the number of Transformer Blocks (TB) in GSRT on the CAVE dataset, including computational complexity and inference time (measured on a $64 \times 64$ patch, averaged over 10 runs).}
\label{tab:hyper_k_tb}
\begin{tabular*}{\textwidth}{@{\extracolsep{\fill}}
  S[table-format=2.0, round-precision=0, detect-weight] 
  S[table-format=1.0, round-precision=0, detect-weight]
  S[table-format=2.3, round-precision=3, detect-weight]   
  S[table-format=2.3, round-precision=3, detect-weight]   
  S[table-format=2.4, round-precision=4, detect-weight]   
  S[table-format=1.2, round-precision=2, detect-weight]   
  S[table-format=2.2, round-precision=2, detect-weight]   
@{}}
\toprule
{$K$} & {TB} & {Params (M) $\downarrow$} & {GFLOPs $\downarrow$} & {PSNR $\uparrow$} & {SAM $\downarrow$} & {Inference Time (ms) $\downarrow$} \\
\midrule
8 & 1 & 10.553 & 18.423 & 49.1465 & 2.1771 &  11.55\\
8 & 2 & 13.308 & 24.169 & 49.1377 & 2.1435 &  13.09\\
8 & 3 & 12.357 & 31.815 & 49.0961 & 2.1064 &  14.39\\
16 & 1 & 10.577 & 18.471 & 49.2673 & 2.1483 &  12.09\\
16 & 2 & 13.332 & 24.218 & 49.1294 & 2.0941 &  13.69\\
16 & 3 & 14.668 & 31.964 & 49.1649 & 2.0712 &  14.95\\
32 & 1 & 10.626 & 18.569 & 49.2741 & 2.1379 &  11.73\\
32 & 2 & 13.380 & 24.315 & 49.3275 & 2.0913 &  12.87\\
32 & 3 & 14.379 & 33.061 & 49.2314 & 2.0772 &  14.75\\
64 & 1 & 10.722 & 19.050 & 49.3013 & 2.1211 &  12.26\\
64 & 2 & 13.477 & 24.509 & 49.5820 & 2.05 &  13.59\\
64 & 3 & 16.232 & 34.255 & 49.6217 & 2.04 &  15.93\\
\bottomrule
\end{tabular*}
\end{table*}

\textbf{Gated Fusion.}
The outputs of the two branches are adaptively blended via a learnable spatial gate:
\begin{equation}
\begin{split}
&[\mathbf{g}_1, \mathbf{g}_2] = \text{Softmax}\big( \text{Conv}_{1\times1}([\mathbf{X}_{\text{spe}}, \mathbf{X}_{\text{spa}}]) \big),\\
&\mathbf{X}_{\text{att}} = \mathbf{g}_1 \odot \mathbf{X}_{\text{spe}} + \mathbf{g}_2 \odot \mathbf{X}_{\text{spa}},
\end{split}
\end{equation}
where $\mathbf{g}_1 + \mathbf{g}_2 = \mathbf{1}$. The gate weights are spatially varying, allowing the network to dynamically adjust the relative importance of spectral and spatial information according to local image content.

\textbf{Transformer Block Update.}
Each GSRT block incorporates the SPGA module and a feed-forward network with residual connections. The block update is:
\begin{equation}
\mathbf{H}_n = \mathbf{H}_{n-1} + \mathbf{X}_{\text{att}} + \text{FFN}\big( \text{LN}(\mathbf{H}_{n-1}+\mathbf{X}_{\text{att}}) \big),
\end{equation}
where $\mathbf{X}_{\text{att}}$ is the output of the SPGA module. Multiple such blocks are cascaded to progressively enhance spectral fidelity, yielding the final output $\hat{\mathbf{Z}}$.

\subsection{Loss Function}
To supervise the network, we employ a combined L1 loss:
\begin{equation}
\mathcal{L} = \alpha \lambda_1 + \beta \lambda_2,
\end{equation}
where $\lambda_1 = \mathcal{L}_1(\hat{\mathbf{Z}}, \mathbf{Z})$ and $\lambda_2 = \mathcal{L}_1(\mathbf{Z}_{\text{init}}, \mathbf{Z})$ denote the L1 losses for the intermediate and final outputs, respectively. $\mathcal{L}_1(\cdot,\cdot)$ denotes the element-wise L1 distance, and $\alpha,\beta$ are balancing weights. In our experiments, we set $\alpha=0.8$, $\beta=0.2$ to prioritize the final reconstruction quality.

\section{Experiments}
\subsection{Experimental Settings}
The proposed ASSR-Net is evaluated on three publicly available datasets: CAVE~\cite{CAVE}, Harvard~\cite{Harvard}, and Gaofen5. For quantitative assessment, five full-reference metrics are employed. These comprise the Peak Signal-to-Noise Ratio (PSNR), Spectral Angle Mapper (SAM), Universal Image Quality Index (UIQI)~\cite{UIQI}, Structural Similarity Index (SSIM)~\cite{SSIM}, and the Erreur Relative Globale Adimensionnelle de Synthèse (ERGAS). Additionally, the no-reference metric QNR is utilized for evaluating real data in the absence of ground truth. We compare ASSR-Net with seven state-of-the-art deep learning-based methods: DHIF-Net~\cite{DHIF}, DSPNet~\cite{DSPNet}, LRTN~\cite{LRTN}, MIMO-SST~\cite{MIMO-SST}, SINet~\cite{SINet}, OTIAS~\cite{OTIAS}, and SRLF~\cite{SRLF}.
\subsection{Training Configuration.}
All models are trained using the Adam optimizer with $\beta_1 = 0.9$ and $\beta_2 = 0.999$, employing a cosine annealing learning rate schedule. The batch size is set to 16, and input patches are of size $64 \times 64$. The training epochs and initial learning rates are dataset-dependent: for CAVE, we train for 1000 epochs with an initial learning rate of $4 \times 10^{-4}$; for Harvard, 200 epochs with $2 \times 10^{-4}$; and for Gaofen5, 2000 epochs with $4 \times 10^{-4}$. For CAVE and Harvard, the degradation pipeline applies a Gaussian blur kernel of size $8 \times 8$ ($\sigma = 3$) followed by $8 \times$ spatial downsampling.All experiments are conducted on a single NVIDIA RTX 4090 GPU.

\subsection{Experimental Results on Simulated Data}
Quantitative comparisons on the CAVE and Harvard datasets are summarized in Table~\ref{tab:comparison}. The proposed ASSR-Net achieves superior performance on both datasets across all evaluation metrics. On the CAVE dataset, ASSR-Net surpasses the second-best method, SRLF, by a margin of 0.2861 dB in PSNR while reducing the SAM by 0.10. These quantitative improvements signify enhanced capabilities in both spatial reconstruction and spectral preservation. Similarly, on the Harvard dataset, ASSR-Net attains a PSNR gain of 0.1019 dB and an SAM reduction of 0.02 compared to SRLF, thereby corroborating its robust generalization capability across diverse scenes and imaging conditions. Qualitative comparisons are provided in Fig.~\ref{fig:compare1} and Fig.~\ref{fig:compare2}. Magnified local regions demonstrate that the proposed method reconstructs sharper textual details and achieves higher spectral fidelity. Furthermore, it exhibits significantly mitigated spectral distortion in both edge regions and homogeneous areas.

\subsection{Experimental Results on Real Data}
For the Gaofen5 dataset, we follow the standard protocol: we spatially downsample the existing LR-HSI and MSI to generate training data. During testing, we input the original LR-HSI and MSI to obtain the HR-HSI. Since the Gaofen5 dataset lacks ground truth HR-HSI, we use the no-reference metric QNR for quantitative evaluation. Table~\ref{tab:qnr_results} shows that ASSR-Net achieves the highest QNR score of 0.9873, outperforming all compared methods. Fig.~\ref{fig:real_data} presents visual comparisons on the Gaofen5 dataset. The results show that ASSR-Net maintains robust performance in real-world conditions. It produces more natural-looking textures and preserves fine spatial details better, while avoiding common artifacts like over-smoothing or spectral contamination.

\sisetup{detect-weight}
\begin{table*}[t]
\centering
\mytablefont
\setlength{\tabcolsep}{0pt}
\renewcommand\arraystretch{\mytablearraystretch}
\caption{Ablation study on CAVE dataset. \checkmark~denotes module included, $\times$~denotes module excluded.}
\label{tab:ablation_study}
\begin{tabular*}{\textwidth}{@{\extracolsep{\fill}}
  *{4}{c} 
  S[table-format=2.4, round-precision=4, detect-weight]
  S[table-format=1.2, round-precision=2, detect-weight]
  S[table-format=1.4, round-precision=4, detect-weight]
  S[table-format=1.4, round-precision=4, detect-weight]
  S[table-format=1.4, round-precision=4, detect-weight]
@{}}
\toprule
DACI & DAE & Fusion & GSRT & {PSNR$\uparrow$} & {SAM$\downarrow$} & {UIQI$\uparrow$} & {SSIM$\uparrow$} & {ERGAS$\downarrow$} \\
\midrule
$\times$ & $\times$ & $\times$ & $\times$ & 47.8014 & 2.4821 & 0.9704 & 0.9932 & 0.5675 \\
\checkmark & $\times$ & $\times$ & $\times$ & 48.1632 & 2.3743 & 0.9712 & 0.9939 & 0.5586 \\
\checkmark & \checkmark & $\times$ & $\times$ & 48.4457 & 2.3176 & 0.9721 & 0.9941 & 0.5412 \\
\checkmark & \checkmark & \checkmark & $\times$ & 48.8363 & 2.2312 & 0.9742 & 0.9952 & 0.5144 \\
$\times$ & \checkmark & \checkmark & \checkmark & 49.2151 & 2.1151 & 0.9746 & 0.9955 & 0.4976 \\
\textbf{\checkmark} & \textbf{\checkmark} & \textbf{\checkmark} & \textbf{\checkmark} & \textbf{49.5820} & \textbf{2.05} & \textbf{0.9769} & \textbf{0.9961} & \textbf{0.4725} \\
\bottomrule
\end{tabular*}
\vspace{-15pt}
\end{table*}

\sisetup{detect-weight}
\begin{table}[t]
\centering
\mytablefont
\setlength{\tabcolsep}{0pt}
\renewcommand\arraystretch{\mytablearraystretch}
\caption{Impact of loss-weight balance ($\alpha \lambda_1 + \beta \lambda_2$) on final performance (CAVE dataset).}
\label{tab:loss_weight}
\begin{tabular*}{\linewidth}{@{\extracolsep{\fill}}
  S[table-format=1.1, round-precision=1, detect-weight]
  S[table-format=1.1, round-precision=1, detect-weight]
  S[table-format=2.4, round-precision=4, detect-weight]
  S[table-format=1.2, round-precision=2, detect-weight]
@{}}
\toprule
{$\alpha$} & {$\beta$} & {PSNR $\uparrow$} & {SAM $\downarrow$}  \\
\midrule
1.0 & 0.0 & 49.1982 & 2.139  \\
0.9 & 0.1 & 49.0507 & 2.07  \\
\textbf{0.8} & \textbf{0.2} & \textbf{49.5820} & \textbf{2.05}  \\
0.7 & 0.3 & 49.3841 & 2.0761  \\
0.6 & 0.4 & 48.9343 & 2.1219  \\
0.5 & 0.5 & 49.0914 & 2.0842  \\
\bottomrule
\end{tabular*}
\end{table}

\begin{table}[t]
\centering
\caption{Classification results (F1 scores, \%) of LR-HSI and the HR-HSI predicted by ASSR-Net on the Houston dataset.}
\label{tab:classification_results}
\mytablefont
\renewcommand\arraystretch{\mytablearraystretch}
\small 
\begin{tabular*}{\columnwidth}{@{\extracolsep{\fill}} lcc}
\toprule
\textbf{Category} & \multicolumn{2}{c}{\textbf{F1 scores(\%)}} \\
\cmidrule(lr){2-3} 
                  & \textbf{LR-HSI} & \textbf{Predicted HR-HSI} \\
\midrule
Healthy grass     & 87.5            & \textbf{92.7}             \\
Stressed grass    & 88.7            & \textbf{94.1}             \\
Synthetic grass   & 92.7            & \textbf{99.9}             \\
Trees             & 83.6            & \textbf{96.5}             \\
Soil              & 97.8            & \textbf{98.8}             \\
Water             & 81.1            & \textbf{97.4}             \\
Residential       & 75.0            & \textbf{91.1}             \\
Commercial        & \textbf{89.6}   & 87.5                      \\ 
Road              & 77.4            & \textbf{83.5}             \\
Highway           & 82.3            & \textbf{88.5}             \\
Railway           & 73.4            & \textbf{91.5}             \\
Parking lot 1     & \textbf{89.4}   & 84.7                      \\ 
Parking lot 2     & 77.9            & \textbf{87.8}             \\
Tennis court      & \textbf{99.4}   & 98.3                      \\ 
Running track     & 91.7            & \textbf{98.2}             \\
\midrule
\textbf{Average F1}            & 85.8            & \textbf{92.7}             \\
\textbf{Average Accuracy (\%)} & 85.3            & \textbf{91.9}             \\
\bottomrule
\end{tabular*}
\end{table}

\subsection{Ablation Studies}
To systematically evaluate the contribution of each core component, we conduct ablation experiments on the CAVE dataset. The results are summarized in Table~\ref{tab:ablation_study}, where we progressively add modules to the baseline (no DACI, no DAE, no Fusion, no GSRT). The baseline achieves a PSNR of 47.80dB and a SAM of 2.48. Adding the DACI module improves PSNR by 0.36dB and reduces SAM by 0.11, demonstrating the benefit of cross‑modal directional interaction. Subsequently incorporating the DAE module further increases PSNR by 0.28dB and lowers SAM by 0.06, validating its ability to capture anisotropic spatial structures. \par
Introducing the Fusion modules  yields a notable gain of 0.39dB in PSNR and a SAM reduction of 0.09, underscoring the importance of multi‑scale adaptive feature aggregation. The final addition of the GSRT module (full model, row 6) brings the most substantial improvement: PSNR rises by 0.75dB and SAM decreases by 0.18 compared to the model without GSRT. This confirms that explicit spectral prior guidance is crucial for correcting spectral contamination.\par
We also examine the necessity of DACI by comparing the full model  with a variant that replaces DACI with a simple addition. Removing DACI causes a PSNR drop of 0.37dB and a SAM increase of 0.06, indicating that directional cross‑modal interaction is essential for optimal spatial–spectral fusion. Overall, the full ASSR-Net configuration achieves a cumulative PSNR gain of 1.78dB (3.72\% relative) and a SAM reduction of 0.43 (17.3\% relative) compared to the baseline. The improvement is substantially larger than the sum of individual module gains, evidencing strong synergy between the directional‑awareness mechanisms and the spectral‑fidelity components in jointly addressing the dual challenges of HSI–MSI fusion.

\begin{figure*}[t]
\centering
\includegraphics[width=\textwidth]{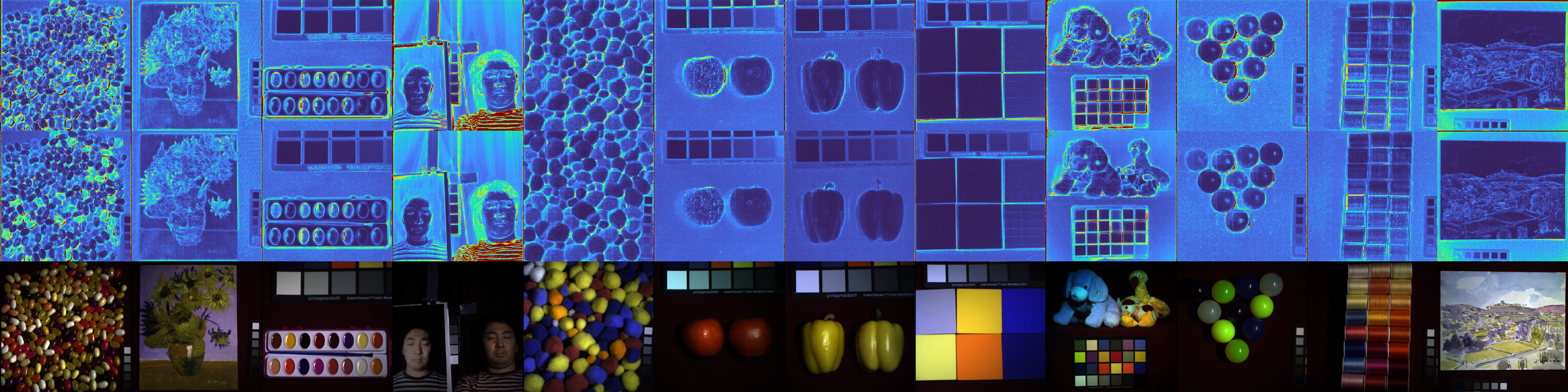}
\caption{Visual comparison of Stage~I and Stage~II outputs. Top row: Spectral error maps (SAM) after Stage~I; middle row: Spectral error maps after Stage~II; bottom row: Pseudo-color RGB images of the final fusion result. Warmer colors in error maps indicate larger spectral deviation. Stage~II significantly reduces errors, especially in complex regions.}
\label{fig:two_stage_visual}
\end{figure*}

\begin{figure*}[t]
\centering
\begin{minipage}{0.75\textwidth}
  \centering
  \includegraphics[width=\linewidth]{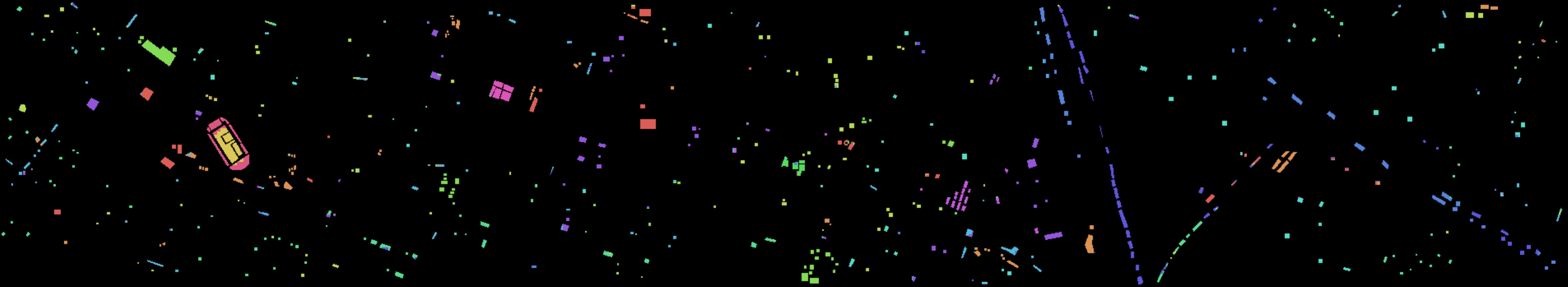}
  (a) Classification result of LR-HSI
  \includegraphics[width=\linewidth]{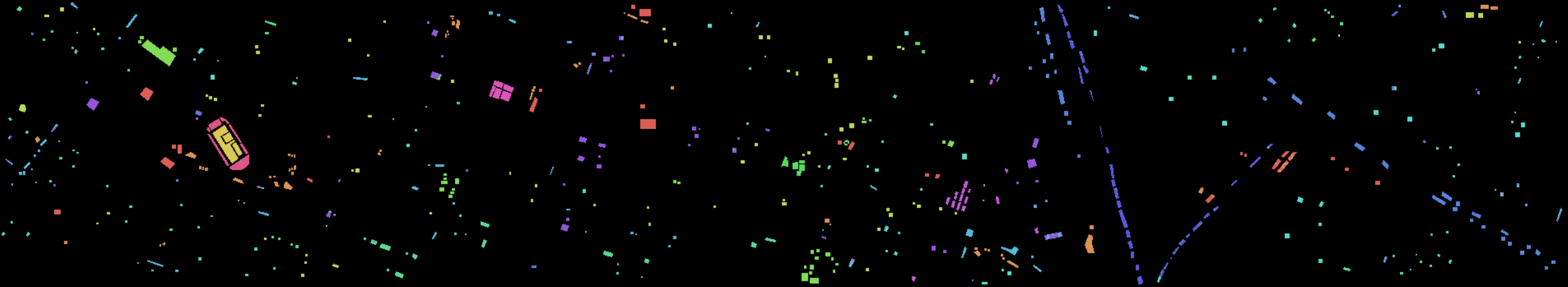}
  \\
  (b) Classification result of the predicted HR-HSI
  \includegraphics[width=\linewidth]{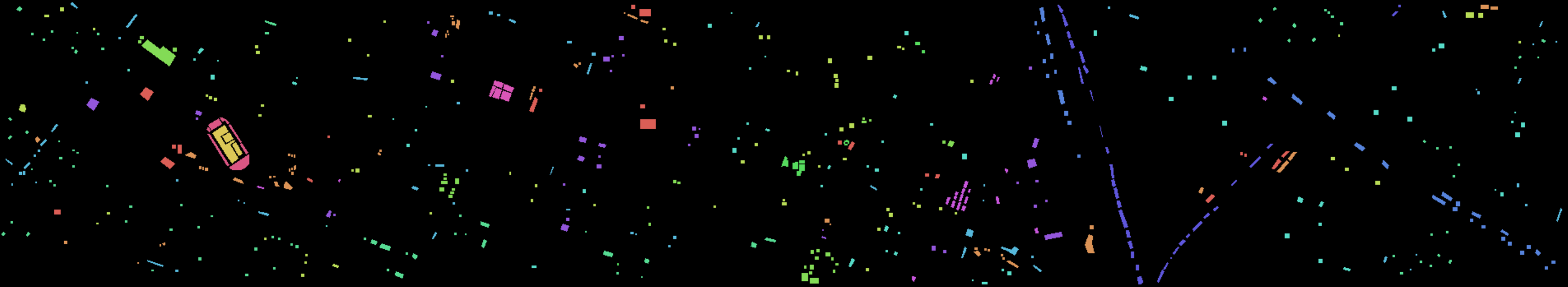}
  \\
  (c) Reference
\end{minipage}
\hfill
\begin{minipage}{0.23\textwidth}
  \centering
  \begin{tikzpicture}[font=\footnotesize]
    \definecolor{color0}{RGB}{0,0,0}       
    \definecolor{color1}{RGB}{221,0,0}     
    \definecolor{color2}{RGB}{255,153,0}   
    \definecolor{color3}{RGB}{255,255,0}   
    \definecolor{color4}{RGB}{153,255,0}   
    \definecolor{color5}{RGB}{0,204,0}     
    \definecolor{color6}{RGB}{0,255,255}   
    \definecolor{color7}{RGB}{0,255,153}   
    \definecolor{color8}{RGB}{0,204,204}   
    \definecolor{color9}{RGB}{0,153,255}   
    \definecolor{color10}{RGB}{0,0,255}    
    \definecolor{color11}{RGB}{102,0,255}  
    \definecolor{color12}{RGB}{153,0,204}  
    \definecolor{color13}{RGB}{204,0,153}  
    \definecolor{color14}{RGB}{255,0,204}  
    \definecolor{color15}{RGB}{255,0,102}  

    \def\legendlist{
      color0/Undefined,
      color1/Healthy grass,
      color2/Stressed grass,
      color3/Synthetic grass,
      color4/Trees,
      color5/Soil,
      color6/Water,
      color7/Residential,
      color8/Commercial,
      color9/Road,
      color10/Highway,
      color11/Railway,
      color12/Parking Lot 1,
      color13/Parking Lot 2,
      color14/Tennis Court,
      color15/Running Track
    }
    \foreach \col/\txt [count=\i from 0] in \legendlist {
      \fill[\col] (0, -\i*0.35) rectangle (0.3, -\i*0.35 - 0.3);
      \node[anchor=west] at (0.4, -\i*0.35 - 0.15) {\txt};
    }
  \end{tikzpicture}
\end{minipage}
\vspace{-5pt} 
\caption{Classification results before and after fusion. (a) Classification result of LR-HSI. (b) Classification result of the predicted HR-HSI. (c) Reference.}
\label{fig:classification}
\end{figure*}
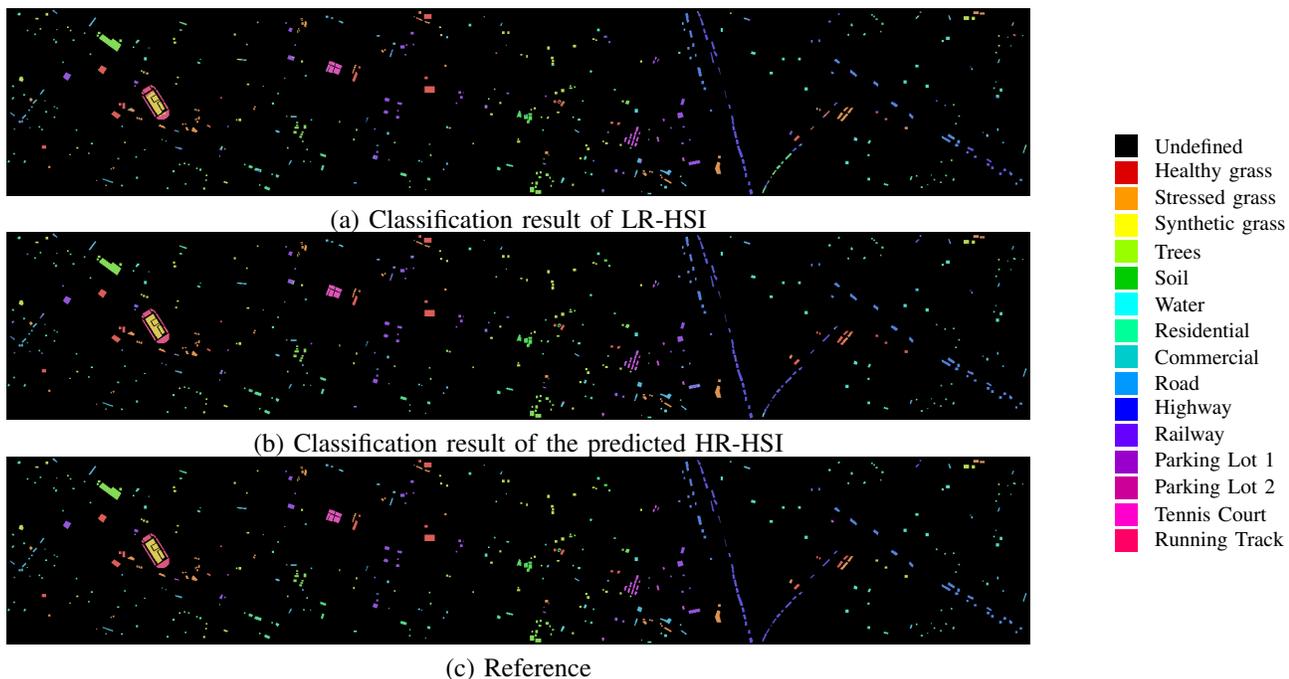

\subsection{Hyperparameter Sensitivity and Complexity Analysis}
\label{sec:hyper_complexity}
We conduct extensive sensitivity analysis on key hyperparameters: the loss weights ($\alpha$, $\beta$), the number of projection directions $K$ in ASSE, and the number of Transformer Blocks (TB) in GSRT. Table~\ref{tab:loss_weight} shows that an optimal balance of $\alpha=0.8$ and $\beta=0.2$ yields the best performance on the CAVE dataset. This ratio indicates that Stage 1 (ASSE) should provide sufficient spatial guidance without excessive spectral distortion, while Stage 2 requires stronger supervision to effectively correct spectral deviations. Table~\ref{tab:hyper_k_tb} presents the performance with varying $K$ and TB. While $K=64$ with $\mathrm{TB}=3$ achieves the highest PSNR, the configuration $K=64$ with $\mathrm{TB}=2$ offers a better trade-off between performance and computational cost, and is selected as our final model.\par
We provide a comprehensive analysis of the model's efficiency. The full ASSR-Net has \textbf{13.477M parameters} and requires \textbf{24.509 GFLOPs} for a single forward pass on a $64\times64\times31$ patch. The average inference time is \textbf{13.59 ms} on an NVIDIA RTX 4090 GPU. For comparison, the first stage (ASSE) alone has 6.856M parameters, 11.124 GFLOPs, and an inference time of 9.43 ms. This demonstrates that the two-stage design, while more sophisticated than single-stage baselines, maintains competitive inference speed due to the efficient design of ASSE. The complexity is comparable to recent advanced fusion methods while delivering superior reconstruction quality, justifying the added computational cost.

\subsection{Effectiveness of the Two-Stage Design}
\begin{figure*}[t]
\centering
\subfloat[Scene A]{\includegraphics[width=0.32\linewidth]{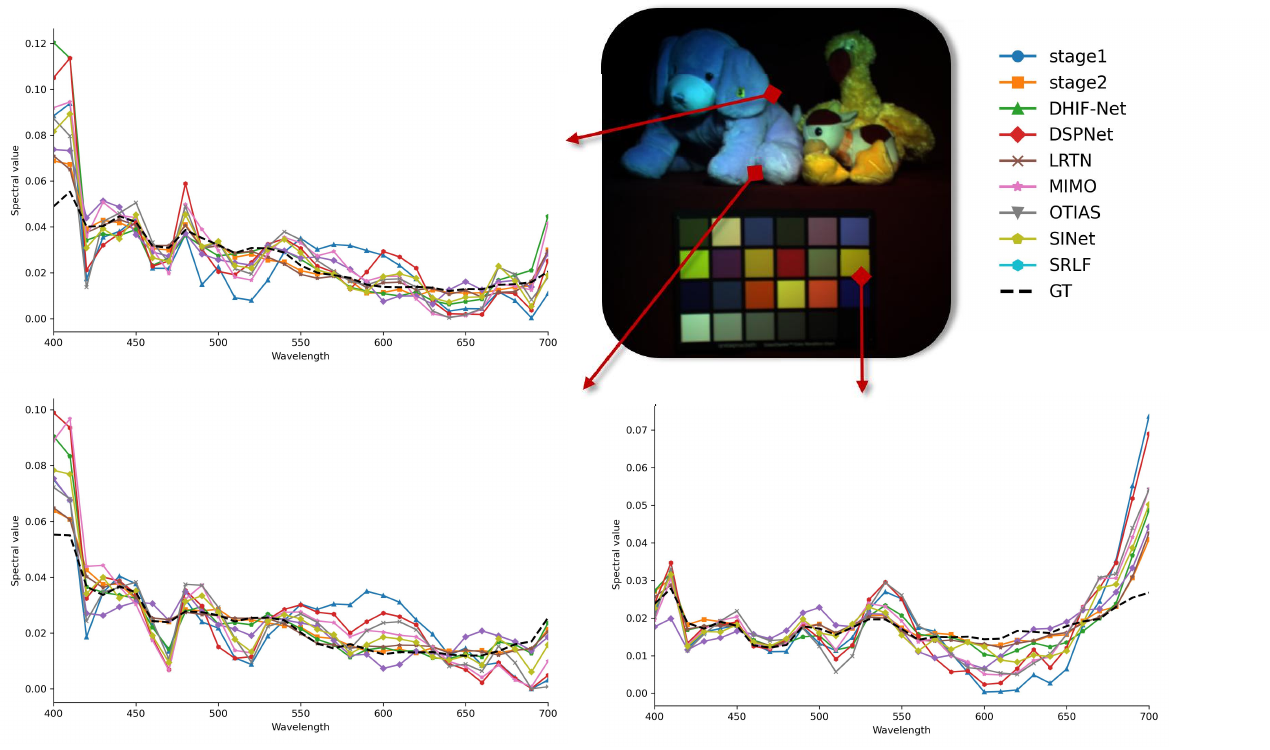}}\label{fig:spec_sceneA}
\hfill
\subfloat[Scene B]{\includegraphics[width=0.32\linewidth]{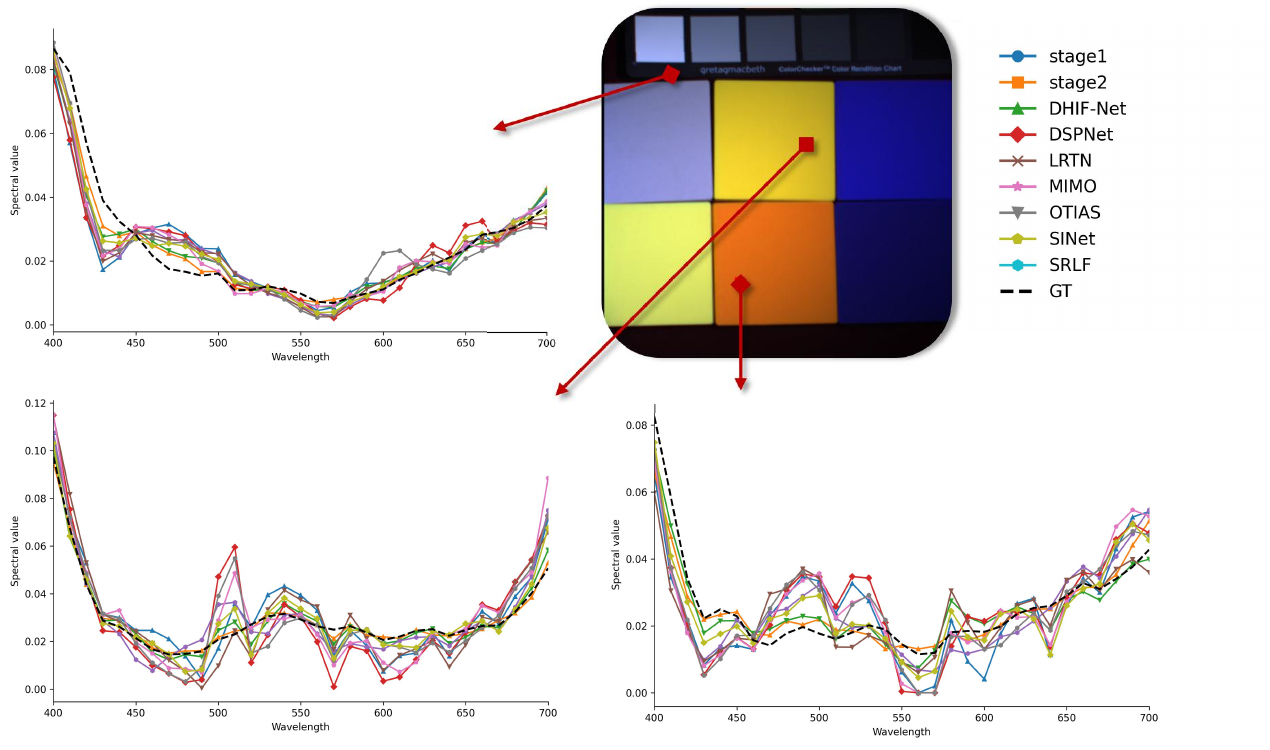}}\label{fig:spec_sceneB}
\subfloat[Scene C]{\includegraphics[width=0.32\linewidth]{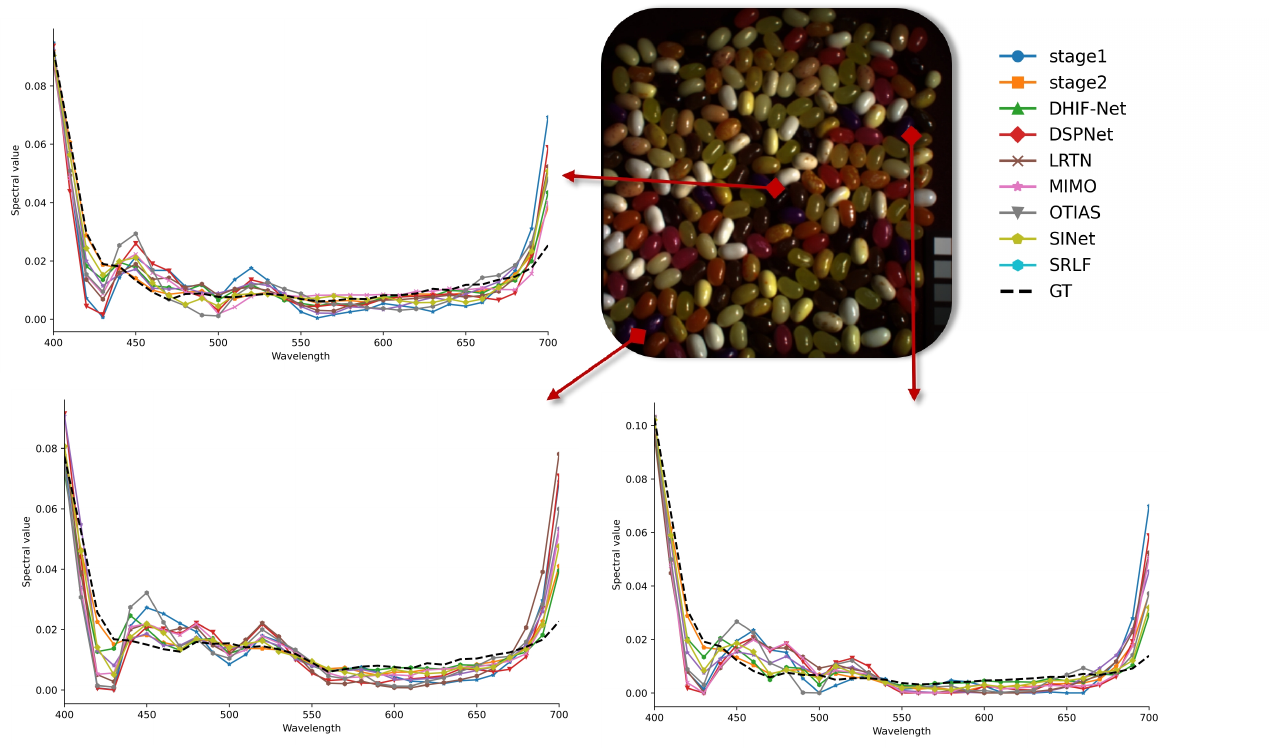}}\label{fig:spec_sceneC}
\caption{Spectral profiles of three points (marked in the RGB images) for three different scenes.Stage~2 consistently reduces spectral deviations compared to Stage~1 and outperforms other methods.}
\vspace{-15pt}
\label{fig:spectral_profiles}
\end{figure*}

To validate the necessity of decoupling spatial enhancement from spectral calibration, we visually compare the outputs of Stage~1 (ASSE) and Stage~2 (HPSC) on representative scenes. As shown in Fig.~\ref{fig:two_stage_visual}, the spectral error maps (SAM) after Stage~1 exhibit noticeable deviations, especially in heterogeneous regions and along object boundaries. After Stage~2, the errors are substantially reduced, demonstrating the efficacy of the GSRT module in correcting spectral distortions. The pseudo-color RGB images confirm that Stage~2 preserves fine spatial details while restoring spectral fidelity.urthermore, we evaluate spectral fidelity at the pixel level by plotting spectral curves of selected points. Fig.~\ref{fig:spectral_profiles} shows three scenes.The curves compare the ground truth, Stage~1 output, Stage~2 output, and several competing methods. Stage~1 often captures the overall shape but exhibits consistent bias across bands, while Stage~2 aligns much closer to the ground truth, particularly in absorption and reflection regions. This confirms that the dedicated spectral calibration step effectively rectifies the spectral contamination introduced during spatial enhancement.

\subsection{The Impact of Fusion on Classification}

To further validate the practical value of the reconstructed HR-HSI, we evaluate its impact on land-cover classification using the Houston dataset \cite{Houston}. Following the protocol in \cite{LRTN}, we generate LR-HSI and MSI from the original HSI (144 bands, $349\times1905$) via spatial and spectral downsampling. The proposed ASSR-Net is trained on degraded pairs and then applied to the original LR-HSI and MSI to produce the HR-HSI. A Support Vector Machine (SVM) classifier with RBF kernel is employed, where optimal parameters ($C$ and $\gamma$) are selected via grid search. 20\% of labeled pixels per class are used for training, and the remaining 80\% for testing. Classification performance is measured by per-class F1 scores and average accuracy.

Table~\ref{tab:classification_results} reports the per-class F1 scores, overall accuracy, and Kappa coefficient for classification on the Houston dataset. Compared with the LR-HSI baseline, the HR-HSI reconstructed by our ASSR-Net achieves substantial improvements across all metrics: the average F1 score increases from 85.8\% to 92.7\% (an improvement of 6.9 percentage points), the overall accuracy rises from 85.3\% to 91.9\% (a gain of 6.6 percentage points), and the Kappa coefficient grows from 0.841 to 0.912 (an increase of 0.071). These results demonstrate that the enhanced spatial resolution and preserved spectral fidelity of our fusion method effectively facilitate more accurate land-cover discrimination.

Figure~\ref{fig:classification} visualizes the classification maps. The result from LR-HSI contains notable noise and misclassifications, particularly in mixed regions and along boundaries. In contrast, the map produced from the predicted HR-HSI is significantly cleaner, with more homogeneous regions and improved consistency with the ground truth. This qualitative comparison further confirms that the HR-HSI reconstructed by our ASSR-Net preserves discriminative spectral information while enhancing spatial details, leading to superior performance in downstream tasks.

\section{Conclusion}
This paper introduces a novel Anisotropic Structure-Aware and Spectrally Recalibrated Network (ASSR-Net), which integrates two principal innovative components. The first component is a Anisotropic Structure-Aware Fusion (ASSE) module, which performs adaptive orientation analysis through learnable geometric transformations, enabling the model to effectively capture the inherent anisotropic spatial structure in remote sensing images. The second is the Global Spectral Recalibration Transformer (GSRT) module, which leverages spectral priors derived from the LR-HSI. It preserves spectral fidelity through a hierarchical guided attention mechanism. Extensive experiments on multiple benchmark datasets demonstrate that the proposed ASSR-Net achieves state-of-the-art performance.



\bibliographystyle{IEEEtran}
\bibliography{main}

\end{document}